%% file: main_arxiv.tex
\documentclass[letterpaper, 10 pt, conference]{ieeeconf}
\IEEEoverridecommandlockouts
\usepackage{graphicx}
\usepackage{amsmath,amssymb}
\usepackage{booktabs}
\usepackage{multirow}
\usepackage{xcolor}
\usepackage{tikz}
\usetikzlibrary{arrows.meta,positioning,calc,fit,backgrounds,shapes.geometric}
\usepackage{url}
\usepackage{cite}
\usepackage{balance}
\usepackage{fontawesome5}   

\newcommand{\ours}{\textsc{Ours}} 
\newcommand{\bfx}[1]{\textbf{#1}}

\title{\LARGE \bf
Learning Social Navigation from Internet Videos\\ in the Policy State Space
}

\author{Jiaming Wang$^{1,*}$, Duc Thang Nguyen$^{1}$, Jizhuo Chen$^{1}$, Volodymyr Shcherbyna$^{2}$, Diwen Liu$^{3}$,\\ Zhengcheng Shen$^{4}$, and Harold Soh$^{1,*}$
\thanks{$^{1}$National University of Singapore, Singapore. $^{2}$Singapore Management University, Singapore. $^{3}$The Chinese University of Hong Kong, Hong Kong SAR, China. $^{4}$Max Planck Institute for Plasma Physics, Germany.}%
\thanks{$^{*}$Corresponding authors: {\tt\small jiaming@comp.nus.edu.sg},\protect\newline {\tt\small harold@comp.nus.edu.sg}}%
}

\begin{document}
\maketitle
\thispagestyle{empty}
\pagestyle{empty}

\begin{abstract}
Training robust social-navigation policies requires simulators with diverse scene layouts, terrain, and human motion, but constructing such environments and specifying pedestrian behavior is costly. We propose an efficient pipeline that converts ordinary monocular walking videos directly into closed-loop social-navigation training environments in the policy's state space. Our key observation is that local social navigation primarily depends on two types of information: where the robot can traverse and how nearby pedestrians move. We therefore represent the static scene as a metric traversability map, which can be rigidly transformed under counterfactual robot motion, while directly replaying the pedestrian trajectories recovered from the video over time. This abstraction allows us to define the forward dynamics directly in the policy's state space and efficiently simulate counterfactual robot states without reconstructing or rendering photorealistic observations. The resulting policy achieves 81.2\% success in the independent Arena benchmark, compared with 75.0\% for the strongest baseline, and succeeds in 19/20 real-robot trials without policy fine-tuning.

\smallskip\noindent\textbf{Project page:} \url{https://jiaming.im/VideoSocNav}
\end{abstract}


\section{Introduction}

Social-navigation policies are commonly trained in simulation because large-scale interaction on real robots is costly and potentially unsafe. These simulators typically use manually constructed environments and pedestrians driven by behavior models such as social force, or learned trajectory predictors~\cite{helbing1995sfm,salzmann2020trajectron++}. This creates two limitations: the diversity of environments is constrained by the cost of constructing layouts, terrain, and obstacles, while the diversity of pedestrian behavior is constrained by the chosen human model.

Recent video-to-simulation and generative scene methods attempt to address the limited diversity of training environments by reconstructing or generating realistic scenes from visual data~\cite{xie2025vid2sim,yoo2026readygo,liu2026urbanverse,wang2026image2sim}. However, these approaches typically require constructing a renderable 3D scene representation---for example through reconstructed geometry, Gaussian splats, or generated assets---and rendering counterfactual observations during simulation, which can be computationally expensive. For social navigation, dynamic pedestrians further require modeling their appearance and motion, adding complexity to simulator construction.

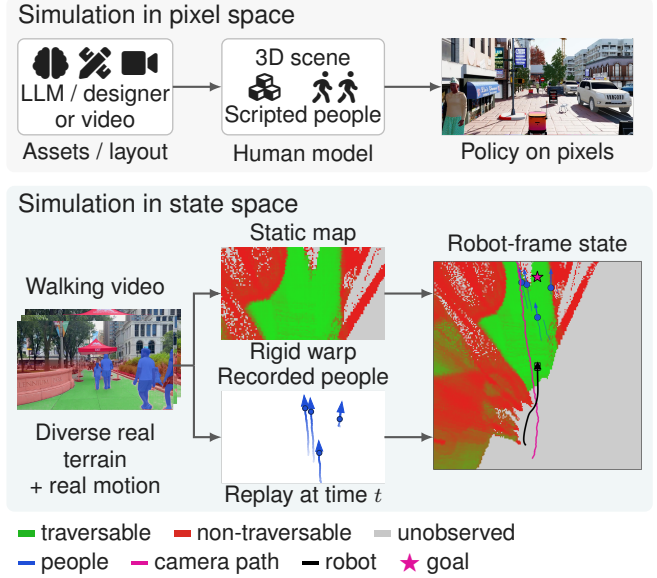
\begin{figure}[t]
\centering
\input{figures/teaser}
\caption{\textbf{From video to social-navigation simulation.}
Ordinary monocular walking videos become closed-loop training environments in the policy's state space. The recorded world is decomposed into a static traversability map, which is rigidly transformed under counterfactual robot motion, and dynamic pedestrian trajectories, which are replayed over time. This lightweight abstraction enables efficient simulation and reinforcement learning without reconstructing or rendering photorealistic scenes.}
\label{fig:teaser}
\vspace{-10px}
\end{figure}

In this work, we build closed-loop social-navigation training environments directly from ordinary monocular walking videos. A social-navigation policy primarily needs to know where it can move and how nearby pedestrians move, so we represent each video directly in the policy's state space, recovering a metric traversability map of the static scene together with pedestrian trajectories from the video (Fig.~\ref{fig:teaser}). During simulation, the map is rigidly transformed with the robot pose, while pedestrians follow their recorded trajectories. This enables reinforcement learning under counterfactual robot actions without reconstructing or rendering the full visual scene.

\begin{figure*}[t]
\centering
\resizebox{\textwidth}{!}{%
\begin{tikzpicture}[
  font=\small,
  lab/.style={align=center, inner sep=1pt, font=\footnotesize},
  stage/.style={font=\large\bfseries, align=center, inner sep=1pt},
  panel/.style={inner sep=0pt, outer sep=0pt, draw=black!45, line width=0.35pt, rounded corners=2.2pt},
  tag/.style={anchor=north west, fill=white, fill opacity=0.8, text opacity=1, inner sep=1.5pt, rounded corners=1.5pt, align=left, font=\normalsize},
  box/.style={draw=black!60, rounded corners=3pt, align=center, inner sep=3pt, fill=white, line width=0.6pt, font=\small},
  arr/.style={-{Latex[length=2.2mm,width=1.8mm]}, line width=1pt, black!70},
  darr/.style={-{Latex[length=2.2mm,width=1.8mm]}, line width=1pt, black!70, dashed},
  alab/.style={font=\normalsize, align=center, inner sep=1.5pt, fill=white},
]
\node[panel] (msk) at (0,0) {\includegraphics[width=3.2cm]{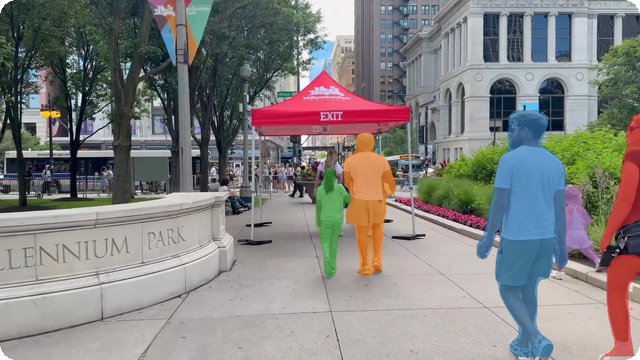}};
\node[panel, below=0.1cm of msk] (trv) {\includegraphics[width=3.2cm]{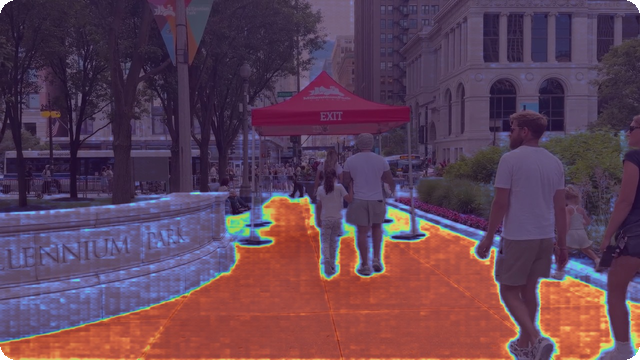}};
\node[panel, below=0.1cm of trv] (dep) {\includegraphics[width=3.2cm]{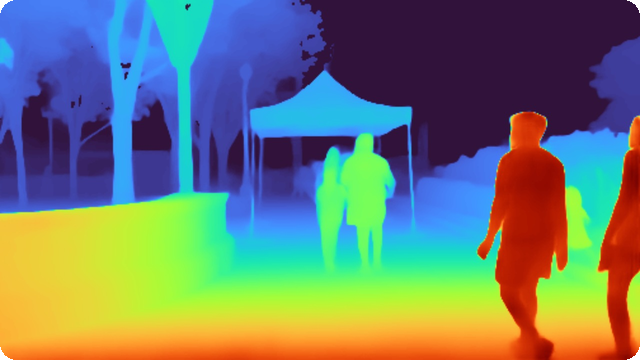}};
\node[tag] at (msk.north west) {people};
\node[tag] at (trv.north west) {traversability};
\node[tag] at (dep.north west) {metric depth + poses};
\coordinate (btop) at (msk.north);
\coordinate (bbot) at (dep.south);
\coordinate (bmid) at ($(btop)!0.5!(bbot)$);
\node[panel, anchor=north east] (v1) at ($(msk.north west)+(-0.85,0)$) {\includegraphics[width=3.2cm]{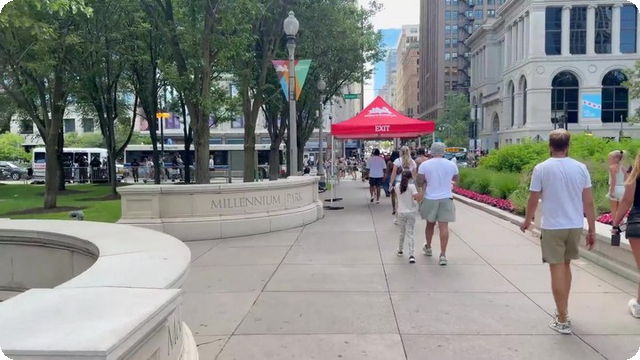}};
\node[panel, below=0.1cm of v1] (v2) {\includegraphics[width=3.2cm]{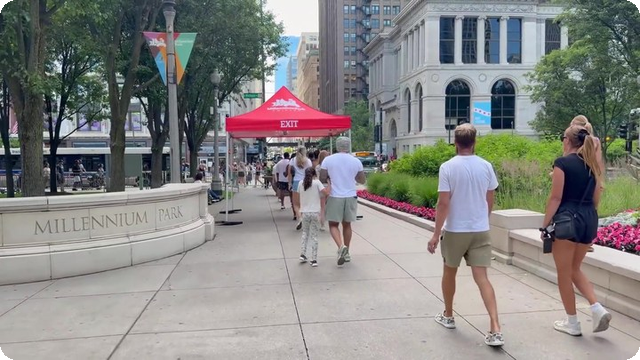}};
\node[panel, below=0.1cm of v2] (v3) {\includegraphics[width=3.2cm]{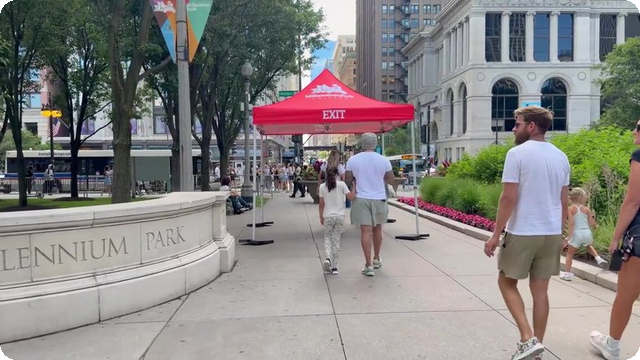}};
\node[tag] at (v1.north west) {$t=4$\,s};
\node[tag] at (v2.north west) {$t=6$\,s};
\node[tag] at (v3.north west) {$t=8$\,s};
\node[panel, anchor=west] (sim) at ($(trv.east |- bmid)+(0.85,0)$) {\includegraphics[height=5.6cm]{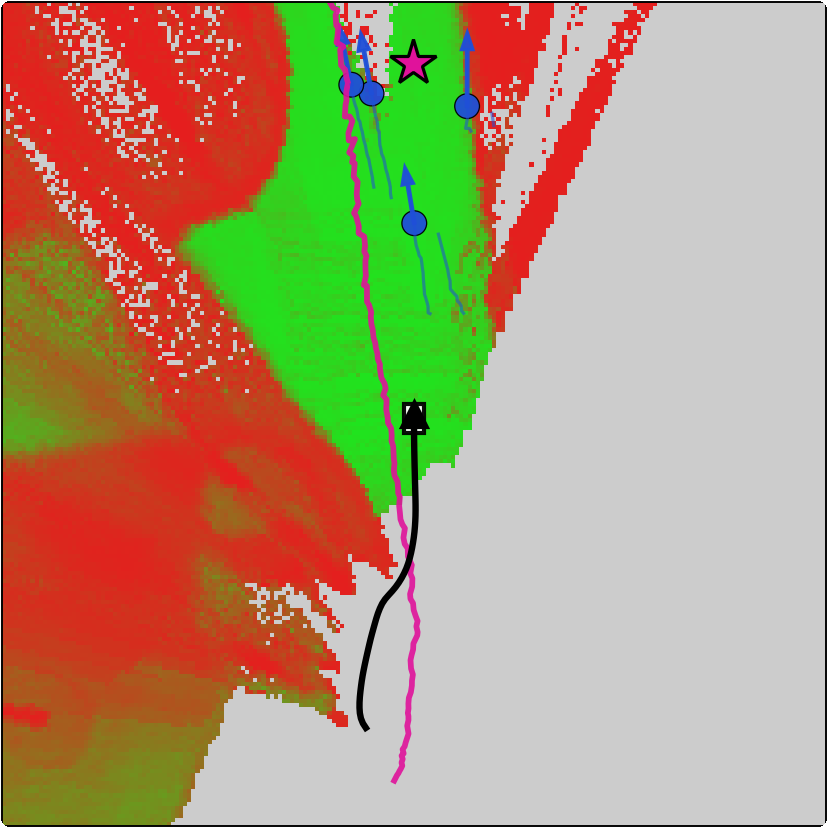}};
\node[tag] at (sim.north west) {robot frame at video time $t$};
\node[tag, anchor=south west] at (sim.south west) {{\color[HTML]{E0129B}\rule[0.5ex]{0.35cm}{1.2pt}} camera path\enskip {\color[HTML]{1F4FD8}\rule[0.5ex]{0.35cm}{1.2pt}} people\\{\color{black}\rule[0.5ex]{0.35cm}{1.2pt}} robot\enskip $\star$ goal};
\coordinate (px) at ($(sim.east)+(0.85,0)$);
\tikzset{chip/.style={draw=black!60, rounded corners=2.5pt, fill=white, inner sep=0pt, minimum width=1.2cm, minimum height=0.85cm, line width=0.5pt, font=\small}}
\node[chip, anchor=north west] (c1) at (px |- btop) {$\mathbf{z}$};
\node[chip, right=0.12cm of c1] (c2) {$B_t$};
\node[chip, right=0.12cm of c2] (c3) {$\mathbf{h}^{1:K}_t$};
\node[chip, right=0.12cm of c3] (c4) {$\mathbf{g}_t^R$};
\node[chip, right=0.12cm of c4] (c5) {$a_{t-1}$};
\node[box, anchor=south east, text width=1.85cm, fill=yellow!22, minimum height=1.6cm] (act) at ($(c3.south |- bbot)+(-0.1,0)$) {actor\\Beta$\to(v_t,\omega_t)$};
\node[box, anchor=south west, text width=1.85cm, fill=gray!15, minimum height=1.6cm] (crit) at ($(c3.south |- bbot)+(0.1,0)$) {critic $V$};
\coordinate (tfx) at ($(c1.west)!0.5!(c5.east)$);
\coordinate (tfnw) at ($(c1.south west)+(0,-0.45)$);
\coordinate (tfse) at ($(c5.south east)+(0,-2.15)$);
\node[box, fill=orange!12, inner sep=0pt, fit=(tfnw)(tfse)] (tf) {transformer\\4 blocks, 203 tokens, $d{=}192$\\RoPE on the cone tokens, 1.6\,M parameters};
\foreach \c in {c1,c2,c3,c4,c5} {\draw[arr, line width=0.6pt] (\c.south) -- (\c.south |- tf.north);}
\draw[arr] (tf.south) -- ++(0,-0.2) -| (act.north);
\draw[arr] (tf.south) -- ++(0,-0.2) -| (crit.north);
\node[panel, anchor=north west] (dep1) at ($(c5.north east)+(0.85,0)$) {\includegraphics[width=3.2cm]{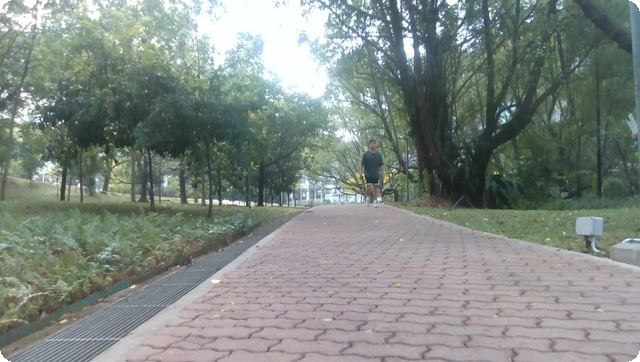}};
\node[panel, anchor=south west] (dep2) at (dep1.west |- bbot) {\includegraphics[width=3.2cm]{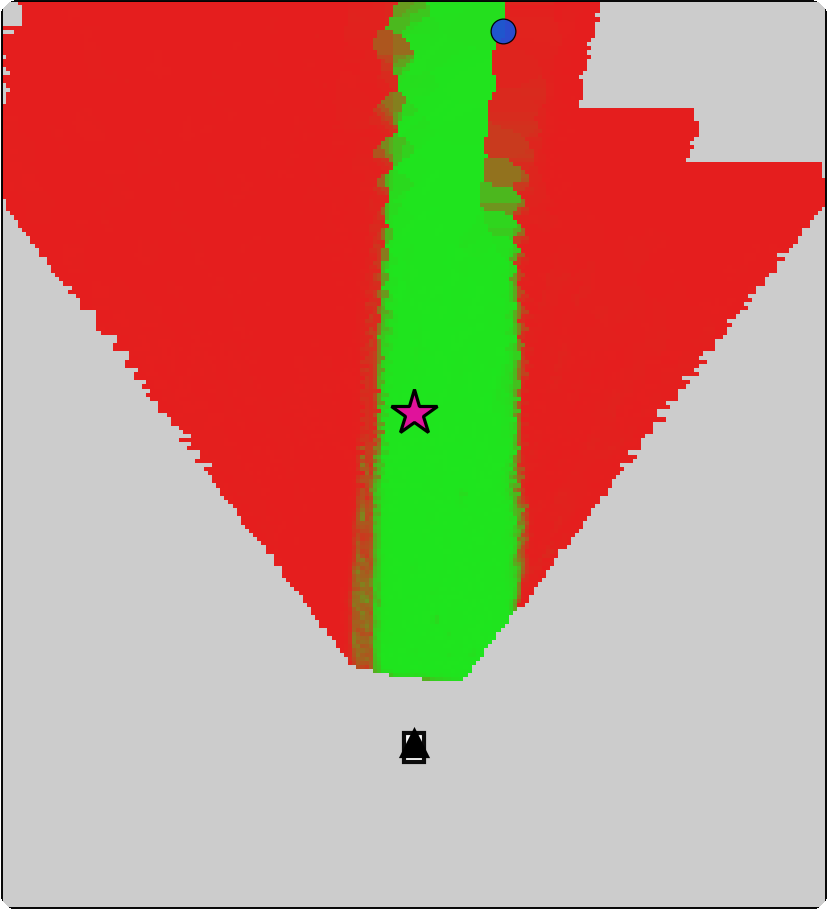}};
\node[tag] at (dep1.north west) {onboard perception};
\node[tag, anchor=south west] at (dep2.south west) {online $B_t$ and tracks};
\draw[arr] (dep1.south) -- (dep2.north);
\path (btop) ++(0,0.12) coordinate (ttop);
\node[stage, anchor=south] at (v1.center |- ttop) {(a) Walking video};
\node[stage, anchor=south] at (msk.center |- ttop) {(b) Data processing};
\node[stage, anchor=south] at (sim.center |- ttop) {(c) Metric replay step};
\node[stage, anchor=south] at (tf.center |- ttop) {(d) Policy};
\node[stage, anchor=south] at (dep1.center |- ttop) {(e) Deployment};
\draw[arr] (v2.east) -- (v2.east -| trv.west);
\draw[arr] (trv.east) -- (sim.west);
\draw[arr] (sim.east |- c3.center) -- (c1.west |- c3.center) node[alab, midway, above=1pt] {$o_t$};
\draw[arr] (act.west) -- (act.west -| sim.east) node[alab, midway, above=1pt] {$a_t$};
\coordinate (dwx) at ($(dep2.west)+(-0.42,0)$);
\coordinate (pin) at (c5.east |- c3.center);
\draw[darr] (dep2.west) -- (dwx) |- (pin);
\node[alab, rotate=90] at ($(dwx)!0.5!(dwx |- pin)$) {online $o_t$};
\end{tikzpicture}}
\caption{\textbf{Overview.}
(a) First-person walking videos are processed by
(b) pretrained perception models to recover a gravity-aligned metric traversability map and time-indexed pedestrian states.
(c) These quantities define the state-space replay simulator: world time follows the recording, while the robot pose evolves independently from the policy commands. The recorded map and pedestrians are transformed into the robot frame at each step, enabling counterfactual motion away from the camera path, collision checking, and closed-loop training without rendering.
(d) The policy encodes the egocentric traversability map, nearby pedestrian states, goal, and previous action with a lightweight transformer and outputs velocity commands.
(e) At deployment, the same policy state is reconstructed online from onboard perception, allowing the video-trained policy to transfer without modification.}
\label{fig:overview}
\vspace{-8px}
\end{figure*}
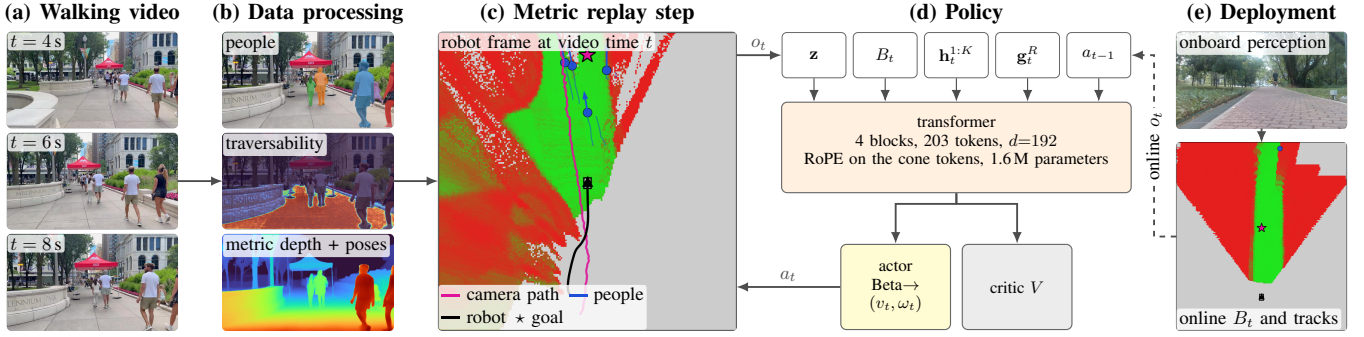

The resulting formulation is lightweight and scalable. We construct 6{,}075 training episodes from 33.8 hours of walking videos spanning diverse cities, terrain, crowd configurations, and pedestrian behaviors. The state-space simulator runs at roughly 1{,}000 environment steps/s on a single RTX~5090, enabling large-scale reinforcement learning. Extensive ablations of the simulator and policy design show, in particular, that directly replaying recorded pedestrian trajectories is sufficient to learn robust social-navigation behavior, without requiring an explicit human behavior model (Sec.~\ref{sec:ablation}). A policy trained solely in these video-derived environments transfers without fine-tuning to the independent Arena simulator, achieving 81.2\% success and outperforming prior state-of-the-art methods. The same policy also transfers to a physical robot, succeeding in 19/20 trials without policy fine-tuning (Sec.~\ref{sec:real}).

Our contributions are threefold:
(1) a state-space simulator that converts ordinary walking videos into closed-loop social-navigation training environments while preserving the recorded pedestrian motion;
(2) a systematic study of policy learning in these video-derived environments, including ablations on training strategy, pedestrian modeling, and simulator design; and
(3) evaluation on unseen videos, the Arena benchmark, and a physical robot, demonstrating generalization and transfer without policy fine-tuning.

The processed dataset, trained policy checkpoints, and code will be released upon publication.

\section{Related Work}

\textbf{Social navigation.}
Social-navigation methods have evolved from reciprocal collision avoidance to learned policies that model pedestrian interactions through reinforcement learning, interaction graphs, and trajectory prediction~\cite{chen2017cadrl,everett2018ga3c,chen2019crowdnav,xie2023drlvo,liu2023attngraph,liu2025height,kumar2025crowdsurfer}. Most are trained with synthetic pedestrian behavior, while real-world datasets such as SCAND, MuSoHu, and SACSoN~\cite{karnan2022scand,nguyen2023musohu,hirose2023sacson} provide more natural interactions but are costly to collect at scale. Rather than proposing another policy architecture, we focus on obtaining scalable closed-loop training from real pedestrian behavior in video.

\textbf{Simulation for social navigation.}
Most social-navigation simulators combine constructed or scanned environments with pedestrians controlled by behavior models such as social force or learned trajectory predictors~\cite{helbing1995sfm,salzmann2020trajectron++}. HuNavSim and Arena, for example, provide configurable reactive crowds~\cite{perez2023hunavsim,11246895}, but the resulting pedestrian behaviors are constrained by the chosen models. In contrast, SocNavBench~\cite{biswas2022socnavbench} replays prerecorded pedestrian trajectories and is designed primarily as an evaluation benchmark, with a fixed set of prepared environments and recorded scenarios rather than diverse training environments. Our approach instead constructs both traversability and pedestrian motion directly from ordinary monocular videos, enabling scalable closed-loop policy training across diverse real-world scenes.

\textbf{Video-to-simulation.}
Recent video-to-simulation methods reconstruct or generate realistic environments from video using Gaussian splats, 3D assets, or generative models~\cite{xie2025vid2sim,chhablani2025embodiedsplat,yoo2026readygo,liu2026urbanverse,wang2026image2sim}. Dynamic social scenes additionally require human representations and motion models, making simulator construction and rendering costly. Other works use web videos as navigation demonstrations~\cite{hirose2024lelan,liu2025citywalker,urbannav2025,vega2026}, but do not provide closed-loop interaction beyond the recorded trajectories. We instead extract the task-relevant state directly from video to build efficient closed-loop social-navigation training environments.

\section{Method}

\subsection{Problem statement and overview}

We consider a wheeled or legged robot equipped with a forward RGB-D camera, tasked with reaching a fixed world-frame goal $\mathbf{g}^W\in\mathbb{R}^2$ while avoiding static obstacles and pedestrians.
At 10\,Hz, the policy $\pi_\theta$ maps an observation $o_t$ to a velocity command
$a_t=(v_t,\omega_t)$. Rather than pixels, the observation contains an egocentric traversability
map $B_t\in[0,1]^{200\times200}$ covering $\pm10$\,m at 0.1\,m resolution, up to $K=20$
pedestrian states $\mathbf{h}^i_t=(x,y,v_x,v_y)$ within 10\,m, the goal $\mathbf{g}_t^R$ expressed in the robot frame, and the previous action.

This representation captures the quantities that determine local navigation outcomes:
which ground is traversable, including terrain and static obstacles, and where nearby people are
and are moving. It also makes video-derived simulation simple: the static scene is a map that
changes under robot motion only by a rigid transform, while pedestrian states can be replayed
directly from the video. Fig.~\ref{fig:overview} shows how monocular walking videos are converted
into this metric state for training, and how the same observation is reconstructed online from
RGB-D and odometry at deployment.

\begin{figure}[t]
\centering
\includegraphics[width=\linewidth]{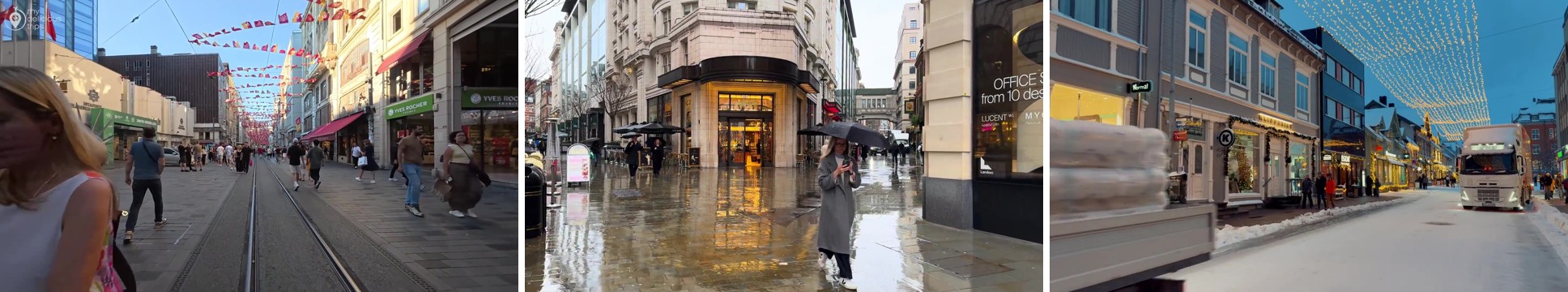}
\caption{Frames from three held-out source videos: an evening crowd on Istiklal Street
(Istanbul), a rainy shopping street (London), and a snow-covered street at dusk (Troms\o{}).
Ordinary first-person walking tours like these are the only input of the pipeline.}
\label{fig:videos}
\vspace{-8px}
\end{figure}

To construct such training environments from monocular video, we must recover all observations
in a common metric frame and keep the world state \emph{temporally consistent}: the map the
robot sees at video time $t$ must not contain ground that the camera only observed later. Each
20\,s, 10\,fps clip (Fig.~\ref{fig:videos}) is therefore processed to recover metric geometry,
traversability, the demonstrator trajectory, and pedestrian motion.

\subsection{Data processing pipeline}
\label{sec:data_process}
\textbf{Metric depth and camera poses.} Depth Anything~3~\cite{lin2025da3} is run once over the
200 frames of a clip (multi-view inference at 504\,px) and jointly estimates \emph{metric} depth,
camera intrinsics, and camera-to-world poses. The recovered camera trajectory provides the
demonstration used for imitation learning in Sec.~\ref{sec:training}.

\textbf{Gravity-aligned traversability.} Navigation is governed by the ground plane rather than
raw 3D geometry. SAM-TP~\cite{wang2025genie} predicts pixel-wise walkability, whose
high-confidence regions are back-projected using the estimated depth. A ground plane fitted by
RANSAC in the best-grounded frames of the clip defines a gravity direction and a metric camera
height (median 1.3\,m); each frame keeps its full 6-DoF pose and is placed in the resulting
gravity-aligned world frame $W$, in which navigation is expressed. The projected camera centers
and optical axes yield the demonstrator poses $(x_t,y_t,\psi_t)$.

\textbf{Causal traversability map.}
For each frame, depth points are transformed into $W$, points inside person boxes are removed, and the remaining measurements are rasterized at 0.1\,m resolution with their traversability scores. Measurements are accumulated only from frames up to the current time using an exponential moving average, producing a map sequence $\{M_t\}$ in which $M_t$ contains measurements only for cells observed by frame $t$. This causality applies only to map accumulation: depth, camera poses, and the ground plane are estimated from the full clip, so individual measurements benefit from batch reconstruction and are more accurate than those available to an online robot. Cells not observed by frame $t$ contain no measurement; in the released data, they are assigned the free prior (traversability~1) in both the policy observation and collision map, making unobserved space indistinguishable from observed free space. We examine this assumption in Sec.~\ref{sec:assumptions}.

\textbf{Pedestrian trajectories.} YOLO11~\cite{yolo11} segmentation and pose estimates detect
visible pedestrians, and BoT-SORT~\cite{aharon2022botsort} associates detections over time
before they are lifted into 3D. Each person is localized on the ground plane by back-projecting
its mask with metric depth, and a constant-velocity Kalman filter per identity smooths the track
and estimates velocity; a track is created after four consecutive in-range detections with a
visible torso and coasts through at most eight missed frames. The resulting time-indexed states
$(x,y,v_x,v_y)_t$ provide the pedestrian dynamics of the simulator.

\textbf{Episode gating and cost.} We retain clips with sufficient camera motion, valid map
reconstruction, and meaningful pedestrian interaction, removing stationary segments, pose
failures, and scenes without nearby people. After filtering, 73\% of candidate clips remain (Table~\ref{tab:data}).
Every stage is feed-forward: converting one 20\,s clip (200 frames) takes 4.5\,min of
wall-clock time on one A100 GPU.

\begin{table}[t]
\caption{Corpus statistics.}
\label{tab:data}
\centering
\footnotesize
\begin{tabular}{lrr}
\toprule
 & Train & Test \\
\midrule
Source videos (distinct cities) & 25 (12) & 19 (17) \\
20\,s episodes extracted & 8{,}147 & 3{,}126 \\
Episodes passing gates & 6{,}075 & 2{,}155 \\
Hours of usable video & 33.8 & 12.0 \\
Pedestrian tracks & 91{,}261 & 29{,}846 \\
Tracks per episode (mean) & 15.0 & 13.8 \\
Pedestrians per frame (mean / max) & 2.5 / 18 & 2.3 / 18 \\
Frames with $\ge1$ pedestrian & 72\% & 68\% \\
Demonstrator path per episode (m) & 23.7 & 21.7 \\
\bottomrule
\end{tabular}
\vspace{-8px}
\end{table}

\subsection{The state-space simulator}
\label{sec:sim}

Each episode contains the causal map sequence $\{M_t\}$, pedestrian states $\{P_t\}$ in the
world frame $W$, and the demonstrator trajectory. The simulator decouples \emph{world time} from
\emph{robot motion}: the recorded scene advances along the video timeline, while the robot pose
$\mathbf{p}_t=(x,y,\psi)$ evolves independently according to the policy. Their interaction is
defined only when the recorded state is transformed into the robot's current frame. This allows
the robot to take actions never observed in the original video without predicting or rendering
new images.

\textbf{World and robot evolution.} At step $t$, the simulator retrieves $M_{f_0+t}$ and $P_{f_0+t}$ from the recording, where $f_0$ is the video frame at which the episode starts, preserving the real scene geometry and the pedestrians' recorded trajectories, speeds, and timing. 
Pedestrians follow their recovered trajectories independently of robot actions; the scope of this approximation is discussed below.
In parallel, the robot pose is integrated from $a_t=(v_t,\omega_t)$ with a constant-twist unicycle model over $\Delta t=0.1$\,s: \begin{equation} 
\begin{aligned} \psi' &= \psi + \omega\Delta t,\\ x' &= x + \tfrac{v}{\omega}(\sin\psi' - \sin\psi), y' = y + \tfrac{v}{\omega}(\cos\psi - \cos\psi'), 
\end{aligned}\label{eq:unicycle} 
\end{equation} 
with the straight-line limit as $|\omega|\!\to\!0$. A static collision is declared when the
20th percentile of the map values under the $0.7\times0.5$\,m footprint rectangle falls below
0.1; pedestrians are represented by 0.25\,m discs. The same robot dynamics are inverted between
successive demonstrator poses to obtain the imitation targets in Sec.~\ref{sec:training}.

\textbf{Observation synthesis.}
The policy observation is produced by sampling $M_{f_0+t}$ on a $200\times200$ grid centered and
oriented at $\mathbf{p}_t$; locations outside the recorded map receive the free prior.
Pedestrian positions and velocities are transformed into the same robot frame, and the $K$
nearest tracks within 10\,m are retained. Because this transformation is defined for any
$\mathbf{p}_t$, the robot can stop, turn, or leave the demonstrator path while observing the
corresponding recorded world state.

\textbf{Start and goal.}
During training, the
start is jittered within a 2\,m forward half-disc around the demonstrator pose and rejected if it
overlaps an obstacle or pedestrian. The world-frame goal $\mathbf{g}^W$ is the first demonstrator position at least 15\,m away, also jittered within 2\,m and required to be traversable. These perturbations prevent the
policy from simply replaying the camera trajectory while keeping the episode close to the
demonstration; evaluation uses the unjittered start and goal.

\subsection{Simulation assumptions}
\label{sec:assumptions}
Counterfactual robot motion introduces two approximations that cannot be resolved from a single recording.

\textbf{Partial static-scene coverage.}
The robot can move into regions never observed by the source video; these cells have no reconstructed traversability measurement and are assigned the free prior. We mitigate this by placing starts and goals near the recorded route, while the progress reward and per-step penalty discourage unnecessary detours. We quantify this using \emph{observed-cell coverage}, the fraction of the robot's local map that is supported by measurements from the video, and \emph{unsupported path}, the fraction of traveled distance for which most of the robot footprint lies on cells never observed anywhere in the clip. On held-out episodes (Table~\ref{tab:coverage}), the policy has similar observed-cell coverage to the recorded walker (21\% vs.\ 23\%) and only moderately more unsupported motion (15.2\% vs.\ 11.2\%). We also train an unknown-aware variant that explicitly marks unobserved cells in the policy input and penalizes the fraction of the robot footprint (with coefficient $-0.05$) lying on them. Relative to the control with the same training budget, this reduces unsupported motion from 14.4\% to 12.7\% but also success from 91.4\% to 87.4\%, suggesting that the original policy does not substantially rely on unsupported space and that conservatively penalizing all unobserved regions can unnecessarily restrict navigation.

\textbf{Non-reactive pedestrian motion.}
Pedestrians follow their recorded trajectories rather than reacting to counterfactual robot actions. Replay nevertheless preserves the real speeds, crossings, crowd configurations, and interaction patterns present in the videos without introducing an additional behavior model. We test this approximation directly in Sec.~\ref{sec:ablation} by training with two reactive alternatives, social force and Trajectron++, and evaluating across all three pedestrian dynamics. Replay training remains robust to reactive pedestrians and performs better overall than either model-based alternative. The same trend holds in the independent Arena benchmark (Sec.~\ref{sec:arena}) and on the real robot (Sec.~\ref{sec:real}), suggesting that preserving diverse recorded human motion is more beneficial than approximating it with the pedestrian models considered here.

\begin{table}[t]
\caption{Observation coverage on held-out episodes: fraction of local-map cells supported by recorded measurements, distance traveled on unobserved space (m/episode and \% of path length), and success rate.}
\label{tab:coverage}
\centering
\small
\setlength{\tabcolsep}{2.5pt}
\begin{tabular}{lcccc}
\toprule
Policy & Observed & \multicolumn{2}{c}{Unsupported path} & Success \\
\cmidrule(lr){3-4}
       & cell (\%) & (m/ep.) & (\%) & (\%)$\uparrow$ \\
\midrule
Recorded walker       & 23 & 1.7 & 11.2 & --   \\
Ours (released)       & 21 & 2.3 & 15.2 & 92.6 \\
Control (30\,M)       & 21 & 2.2 & 14.4 & 91.4 \\
Unknown-aware (30\,M) & 21 & 1.8 & 12.7 & 87.4 \\
\bottomrule
\end{tabular}
\vspace{-8px}
\end{table}

\subsection{Policy}
\label{sec:policy}

The policy operates directly on the state representation (Fig.~\ref{fig:overview}d). We
discretize the 10\,m neighborhood around the robot into 180 angular cones of $2^\circ$ and
encode the near-to-far traversability values within each cone (at most 183 cells, padded) as a
single token through one linear layer. The map tokens $B_t$ are combined with up to $K=20$ pedestrian tokens
$\mathbf{h}^i_t=(x,y,v_x,v_y)$, a goal token $\mathbf{g}_t^R$, a previous-action token $a_{t-1}$,
and a readout token $\mathbf{z}$.

A lightweight 4-block transformer (203 tokens of width 192; each block applies RMSNorm,
attention with 6 query heads and one shared key/value head, and a SiLU feed-forward layer of
width 768; 1.6\,M parameters) processes these tokens. Rotary position
encoding~\cite{su2024rope} is applied only to the ordered angular map tokens, while pedestrian
tokens remain permutation-invariant. A learned readout token aggregates the resulting state
representation and is shared by the actor and critic heads. The actor parameterizes independent
Beta distributions $u_v,u_\omega\in[0,1]$, which are affinely scaled to the action bounds as
$v=2.6u_v-1.3$\,m/s and $\omega=4u_\omega-2$\,rad/s. The bounded Beta
parameterization~\cite{chou2017beta} avoids action clipping during PPO, while the critic predicts
the state value from the same readout token.

\subsection{Training with reinforcement learning}
\label{sec:training}

We train the policy from scratch with PPO~\cite{schulman2017ppo} in the state-space simulator of Sec.~\ref{sec:sim}. Each episode uses the counterfactual starts and closed-loop robot dynamics described above, exposing the policy to states and actions beyond the recorded camera trajectory. We also experimented with initializing the policy by imitating the recovered camera motion before PPO, but controlled ablations show no benefit and slightly worse final performance; we therefore use PPO from scratch throughout (Sec.~\ref{sec:ablation}).

The reward primarily encourages goal progress while penalizing collisions and unnecessary motion. Reaching within 0.3\,m of the goal gives $+5$, while a collision with a pedestrian or non-traversable ground terminates the episode with $-5$. At all other steps,
\begin{equation}
\begin{aligned}
r_t ={}& 0.5(d_{t-1}-d_t)-0.01 -0.05\max(0,-v_t) \\
       &-0.005\max(0,|\omega_t-\omega_{t-1}|-0.7),
\end{aligned}
\label{eq:reward}
\end{equation}
where $d_t=\|\mathbf{p}^{xy}_t-\mathbf{g}^W\|_2$ is the planar distance to the goal. The first term rewards progress, the second is a per-step slack penalty, and the third discourages backward motion. The last term encourages smooth motion by penalizing abrupt changes in yaw rate beyond a 0.7\,rad/s deadband. Episodes terminate on success or collision and truncate when the video ends or after 180 steps.

We train with 32 parallel environments using PPO clipping 0.2, $\gamma=0.99$, GAE $\lambda=0.95$, and a learning rate annealed from $5\times10^{-5}$ to $10^{-5}$. A 100\,M-step run takes 26\,h on one RTX~5090, with the state-space simulator executing roughly 1{,}000 environment steps/s.

\section{Experiments}
\label{sec:exp}

We evaluate whether state-space simulation produces a navigation policy that (i) generalizes to
unseen walking videos, (ii) benefits from the proposed simulator and training design,
(iii) transfers to an independent simulator and competes with social-navigation policies
trained conventionally, and (iv) transfers to a physical robot without policy fine-tuning.

\subsection{Data and held-out evaluation}
\label{sec:data}
We collect 44 public 4K first-person walking-tour videos spanning 29 cities
(Fig.~\ref{fig:videos}) and process them as described in Sec.~\ref{sec:data_process}. Training
and test videos are disjoint, with no city shared between the two splits
(Table~\ref{tab:data}). After filtering, the corpus contains 6{,}075 training and 2{,}155 test
episodes, with roughly 121k pedestrian tracks in total. This provides far more natural social encounters than robot-collected
datasets~\cite{karnan2022scand,nguyen2023musohu}, while retaining the layouts, terrain,
and crowd densities of real streets.

\begin{table}[t]
\caption{Closed-loop evaluation on the held-out episodes in the state-space simulator.}
\label{tab:offline}
\centering
\begin{tabular}{lr}
\toprule
Success rate (95\% Wilson interval) & \bfx{92.6\%} [91.4, 93.7] \\
Pedestrian collision & 3.7\% \\
Static-obstacle collision & 1.9\% \\
Timeout (18\,s) & 1.8\% \\
SPL & 0.860 \\
Path efficiency (shortest / traveled) & 0.928 \\
Time to goal (s) & 12.2 \\
Mean speed (m/s) & 1.28 \\
Time in personal zone ($<1.2$\,m) per episode (s) & 0.24 \\
\bottomrule
\end{tabular}
\end{table}

We first evaluate closed-loop in the state-space simulator on every held-out episode: the robot starts
at the demonstrator's pose at $t=2$\,s and must reach the first demonstrator position at least
15\,m ahead, among the recorded people. Table~\ref{tab:offline} reports the standard success, collision,
SPL~\cite{anderson2018spl, wang2024probable} and comfort metrics. The policy succeeds
in 92.6\% of episodes from 19 videos it has never seen, with paths within 8\% of the
straight-line length, and spends a quarter of a second per episode inside anyone's personal space.
Half the remaining failures are pedestrian collisions, and they concentrate in a small number
of particularly complex scenes with dense dynamic traffic, including an autumn shopping street
in Zurich with trams and cyclists (27 of 159 failures), an evening walk in Paris (19) and
evening walks in Barcelona and Istanbul. For reference, replaying the recorded camera trajectory instead of the robot satisfies the simulator's collision criteria in only 72\% of episodes, with 21\% pedestrian and 7\% static collisions.

\subsection{Ablations on held-out videos}
\label{sec:ablation}

\begin{table}[t]
\caption{Ablations on held-out episodes. Succ. is success rate, Ped.\ coll.\ and Stat.\ coll.\ are pedestrian and static-obstacle collision rates, and Pers.\ is time per episode within 1.2\,m of a pedestrian. Brackets give 95\% Wilson confidence intervals. All variants are trained for 30\,M steps.}
\label{tab:ablation}
\centering
\normalsize
\setlength{\tabcolsep}{3pt}
\resizebox{\columnwidth}{!}{%
\begin{tabular}{lccccc}
\toprule
Variant & Succ.$\uparrow$ [95\% CI] & SPL$\uparrow$ &
Ped.\ coll.$\downarrow$ & Stat.\ coll.$\downarrow$ & Pers.\ (s)$\downarrow$ \\
\midrule
Control  & 91.4 [90.1, 92.5] & 0.858 & 4.4 & 1.5 & 0.19 \\
\midrule
IL only                           & 55.3 [53.2, 57.4] & 0.548 & 32.1 & 11.2 & 0.46 \\
PPO from IL initialization                       & 90.0 [88.6, 91.2] & 0.859 & 5.4 & 1.2 & 0.45 \\
No spawn jitter                    & 90.4 [89.1, 91.6] & 0.860 & 5.4 & 1.6 & 0.58 \\
90$^\circ$ FOV map                    & 88.9 [87.5, 90.1] & 0.846 & 4.5 & 1.3 & 0.09 \\
Proxemics reward                   & 85.0 [83.4, 86.5] & 0.823 & 11.5 & 2.1 & 0.42 \\
Square-patch tokens                & 89.7 [88.3, 90.9] & 0.855 & 5.4 & 1.9 & 0.39 \\
25\% training videos               & 81.8 [80.1, 83.3] & 0.791 & 13.5 & 2.4 & 0.28 \\
50\% training videos               & 88.2 [86.8, 89.5] & 0.834 & 5.0 & 0.9 & 0.24 \\
Social-force pedestrians           & 75.6 [73.7, 77.4] & 0.738 & 21.1 & 1.6 & 0.44 \\
Trajectron++ pedestrians           & 90.0 [88.7, 91.2] & 0.850 & 5.7 & 1.8 & 0.34 \\
\midrule
\multicolumn{6}{l}{\emph{Same episodes, pedestrians driven by social-force model}} \\
Control          & 93.4 [92.3, 94.4] & 0.872 & 1.8 & 1.5 & 0.18 \\
Social-force pedestrians           & 88.7 [87.3, 89.9] & 0.859 & 7.6 & 1.4 & 0.59 \\
Trajectron++ pedestrians           & 93.3 [92.1, 94.3] & 0.878 & 2.3 & 1.7 & 0.32 \\
\midrule
\multicolumn{6}{l}{\emph{Same episodes, pedestrians driven by Trajectron++ model}} \\
Control          & 91.9 [90.7, 93.0] & 0.866 & 4.1 & 1.5 & 0.20 \\
Social-force pedestrians           & 75.5 [73.6, 77.2] & 0.738 & 21.8 & 1.3 & 0.46 \\
Trajectron++ pedestrians           & 91.3 [90.0, 92.4] & 0.870 & 5.2 & 1.6 & 0.37 \\
\bottomrule
\end{tabular}}
\end{table}

We study the main training and simulator design choices using the same 2{,}155 held-out episodes, with each PPO variant trained for 30\,M environment steps. Besides the PPO-from-scratch control, we evaluate imitation learning (IL) of the recovered demonstrator actions and PPO initialized from the IL checkpoint. We ablate spawn jitter; restrict the map input to the robot's current $90^\circ$ forward field of view; add a graded proxemics penalty within 1.2\,m of pedestrians ($-0.1$ within 0.45\,m, with magnitude decreasing linearly to $0$ at 1.2\,m); and replace the 180 angular-cone map tokens with $10\times10$ non-overlapping $2\times2$\,m square patches. We also vary the amount of training video and replace replayed pedestrians with reactive social-force (SF) or Trajectron++ dynamics.

\begin{table*}[t]
\caption{Arena/Isaac Sim results on 160 scenarios.
Personal and intimate denote seconds per episode within 1.2\,m and 0.45\,m of a pedestrian, respectively.}
\label{tab:arena}
\centering
\small
\setlength{\tabcolsep}{4pt}
\resizebox{\textwidth}{!}{%
\begin{tabular}{l cc ccc ccc cc}
\toprule
& \multicolumn{2}{c}{Inputs}
& \multicolumn{3}{c}{Outcome (\%)}
& \multicolumn{3}{c}{Efficiency}
& \multicolumn{2}{c}{Social} \\
\cmidrule(lr){2-3}
\cmidrule(lr){4-6}
\cmidrule(lr){7-9}
\cmidrule(lr){10-11}
Method & LiDAR & Plan
& Success [95\% CI]$\uparrow$
& Coll.$\downarrow$ (ped/wall)
& Timeout$\downarrow$
& Time (s)$\downarrow$ & SPL$\uparrow$ & $\bar v$ (m/s)
& Personal (s)$\downarrow$ & Intimate (s)$\downarrow$ \\
\midrule
CrowdSurfer~\cite{kumar2025crowdsurfer}
& \checkmark & \checkmark
& 42.5 [35.1, 50.2] & 50.0 (47.5/2.5) & 7.5
& 46.2 & 0.418 & 0.49
& 19.9 & 1.9 \\
AttnGraph~\cite{liu2023attngraph}
&  & \checkmark
& 35.0 [28.0, 42.7] & 45.6 (11.2/34.4) & 19.4
& 52.0 & 0.299 & 0.47
& 10.3 & \bfx{1.2} \\
HEIGHT~\cite{liu2025height}
& \checkmark & \checkmark
& 75.0 [67.8, 81.1] & 25.0 (19.4/5.6) & \bfx{0.0}
& 46.4 & 0.714 & 0.47
& 11.8 & 3.5 \\
\ours{} (SF peds)
& &
& 58.1 [50.4, 65.5] & 41.9 (39.4/2.5) & \bfx{0.0}
& \bfx{42.6} & 0.559 & 0.49
& \bfx{8.4} & 1.8 \\
\ours{} (Trajectron++ peds)
& &
& 70.0 [62.5, 76.6] & 25.0 (20.0/5.0) & 5.0
& 51.9 & 0.599 & 0.48
& 12.2 & 2.3 \\
\ours{}
& &
& \bfx{81.2} [74.5, 86.5] & \bfx{15.0} (13.8/1.2) & 3.8
& 46.2 & \bfx{0.747} & 0.49
& 10.6 & 2.1 \\
\bottomrule
\end{tabular}}
\end{table*}

\textbf{Replay is an effective approximation of pedestrian behavior.}
A central approximation of our simulator is to replay recorded pedestrian
motion rather than synthesize how pedestrians would react to counterfactual
robot actions. We therefore compare replay against two reactive alternatives.
For social force~\cite{helbing1995sfm}, we use the standard interaction model
while setting each pedestrian's preferred motion toward its recorded
trajectory, so pedestrians follow the observation when unperturbed but react
to the robot and one another. As a learned alternative, we train
Trajectron++~\cite{salzmann2020trajectron++} from scratch on our training
videos, since the released ETH/UCY model transfers poorly to our
walking-tour data. Pedestrian tracks and the camera carrier are treated as
agents in a common world frame and resampled at 0.4\,s; the model uses up to
2.8\,s of history to predict 1.2\,s of future motion. We train on 5{,}629
scenes containing 82{,}880 pedestrian tracks, with no test-video data used. During simulation, Trajectron++ is
conditioned on the simulated histories of both pedestrians and the robot.

Across all three evaluation dynamics, the replay-trained policy is consistently robust, indicating that training directly on recorded pedestrian motion generalizes well even when pedestrians become reactive at test time. Training with social-force pedestrians performs substantially worse, while Trajectron++ is competitive with replay but offers no clear benefit despite requiring an additional learned model and running more slowly: replay runs at roughly 1{,}000 steps/s, compared with about 800 steps/s for Trajectron++. The same trend also holds in the independent Arena benchmark (Sec.~\ref{sec:arena}) and real robot experiment (Sec.~\ref{sec:real}), where the replay-trained policy again performs best. Overall, these results suggest that direct trajectory replay provides a simple, efficient, and sufficiently diverse training distribution without requiring an explicit reactive pedestrian model.

\textbf{More video improves social navigation.}
Performance increases consistently with the amount of training video:
using 25\%, 50\%, and 100\% of the videos gives 81.8\%, 88.2\%, and
91.4\% success, respectively. The clearest change is in pedestrian
collisions, which fall from 13.5\% to 5.0\% and then 4.4\%, while static
collision rates remain low and show no comparable monotonic trend.
This suggests that scaling the video collection is particularly valuable
for exposing the policy to more diverse crowd configurations and
pedestrian behaviors.

\textbf{Closed-loop interaction is essential.}
Supervised imitation of the recovered demonstrator motion reaches only 55.3\%
success, compared with 91.4\% for PPO from scratch. Initializing PPO from
the imitation policy also provides no benefit (90.0\%). Thus, the value of
video is not merely as a source of demonstrations: converting it into an
interactive environment allows the policy to encounter off-trajectory
states, failures, and recovery actions that are absent from imitation.

\textbf{Other design choices.}
The remaining choices have comparatively modest effects. Restricting the map input to the current $90^\circ$ forward field of view reduces success to 88.9\%, while replacing angular-cone tokens with square patches
gives 89.7\%. In contrast, adding a graded proxemics reward substantially degrades performance to 85.0\% with increased pedestrian collisions. Removing spawn jitter changes success by only
1.0 percentage point (90.4\% versus 91.4\%), indicating that the method does not
depend strongly on this augmentation.

\subsection{Arena benchmark}
\label{sec:arena}

\begin{figure}[t]
\centering
\includegraphics[width=0.98\linewidth]{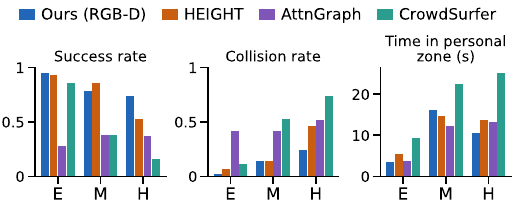}
\caption{Arena results by crowd density: E, M and H are the easy (0--4 pedestrians), medium (5--9) and hard (10--14) scenarios. The advantage of our video-trained policy grows with density.}
\label{fig:arena_diff}
\end{figure}

\textbf{Setup.}
We test transfer beyond the state-space simulator using
Arena~\cite{11246895}, with photorealistic Isaac Sim
environments and reactive social-force pedestrians. Four UrbanVerse
street scenes each contain 40 scenarios: ten frontal encounters,
five with pedestrians moving in the robot's direction, five crossing
encounters, and twenty mixed scenarios. Crowd sizes range from
0--4 pedestrians (\emph{easy}) to 5--9 (\emph{medium}) and
10--14 (\emph{hard}).
A Clearpath Jackal with an OAK-D RGB-D camera navigates between goals
18--22\,m apart. Success requires reaching within 1\,m of the goal
within 180\,s; any wall or pedestrian contact terminates the episode.
Our policy runs at 10\,Hz without fine-tuning.

\textbf{Baselines and inputs.}
We evaluate three learning-based planners:
HEIGHT~\cite{liu2025height},
AttnGraph~\cite{liu2023attngraph}, and
CrowdSurfer~\cite{kumar2025crowdsurfer}.
We use the official checkpoints and Arena's integration of all three methods.
All three follow subgoals from a Nav2 global path; HEIGHT and CrowdSurfer also receive 360$^\circ$ LiDAR, whereas AttnGraph uses no LiDAR. Our policy instead uses RGB-D frames and odometry to construct
its traversability map, without LiDAR, a prebuilt map, or a global
planner. Since the simulator does not provide the LiDAR intensities
required by the baselines' LiDAR-based pedestrian detector, and to isolate navigation performance
from detection quality, we provide all methods with ground-truth pedestrian
positions and estimate velocities using the Kalman tracker.

We train each of the three variants of our policy for 200\,M environment steps and, for each variant, select the checkpoint with the highest rollout reward for evaluation.

\textbf{Results.}
Trained solely in video-derived replay environments, our policy transfers to Arena's reactive crowds and achieves the highest success rate (81.2\%) and SPL (0.747), despite using no LiDAR, prebuilt map, or global planner (Table~\ref{tab:arena}). Compared with HEIGHT, the strongest baseline, it improves success by 6.2 percentage points and reduces collisions from 25.0\% to 15.0\%, including both pedestrian contacts (13.8\% vs.\ 19.4\%) and wall contacts (1.2\% vs.\ 5.6\%).

The advantage becomes larger as crowd density increases. On hard scenarios, our policy reaches 74.2\% success, compared with 53.2\% for HEIGHT, 37.1\% for AttnGraph, and 16.1\% for CrowdSurfer (Fig.~\ref{fig:arena_diff}). Our policy does not simply maximize clearance from pedestrians: it spends 10.6\,s and 2.1\,s per episode in the personal and intimate zones, respectively, not the lowest values among all methods but below HEIGHT's 11.8\,s and 3.5\,s, while achieving substantially higher success and fewer collisions. Together, these results suggest that replay training teaches the policy to maintain progress through dense crowds, including brief close but non-colliding passes that are common in the recorded walking videos, rather than becoming overly conservative or simply relying on unsafe proximity.

Arena also provides an out-of-distribution test of the pedestrian-motion approximation studied in Sec.~\ref{sec:ablation}. Policies trained in the same video-derived scenes but with SF pedestrians and Trajectron++ pedestrians reach only 58.1\% and 70.0\% success, respectively, compared with 81.2\% for direct replay. The Trajectron++ result is particularly informative: although it is trained on the same extracted pedestrian trajectories and performs nearly on par with replay on our held-out video environments, its advantage does not carry over to Arena. One plausible explanation is that a learned trajectory model necessarily approximates the empirical behavior distribution and may smooth rare interaction patterns or accumulate model error during repeated rollout. Direct replay instead preserves the original speeds, crossings, yielding behavior, and crowd interactions without this additional modeling bottleneck. Combined with the ablations, replay is thus not only simpler and faster but also transfers more robustly to pedestrian dynamics not seen during training.

\begin{figure}[t]
\centering
\includegraphics[height=2.1cm]{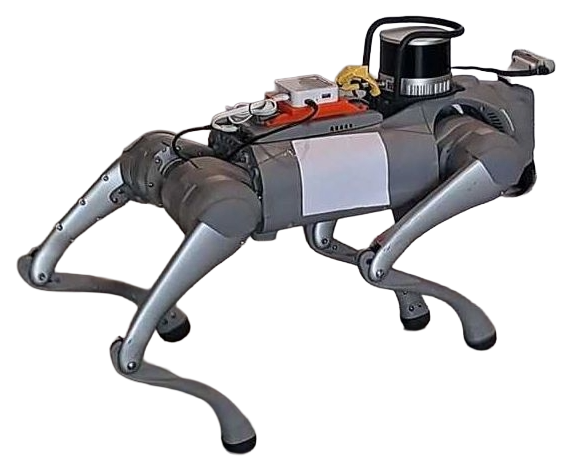}\hfill
\includegraphics[height=2.1cm]{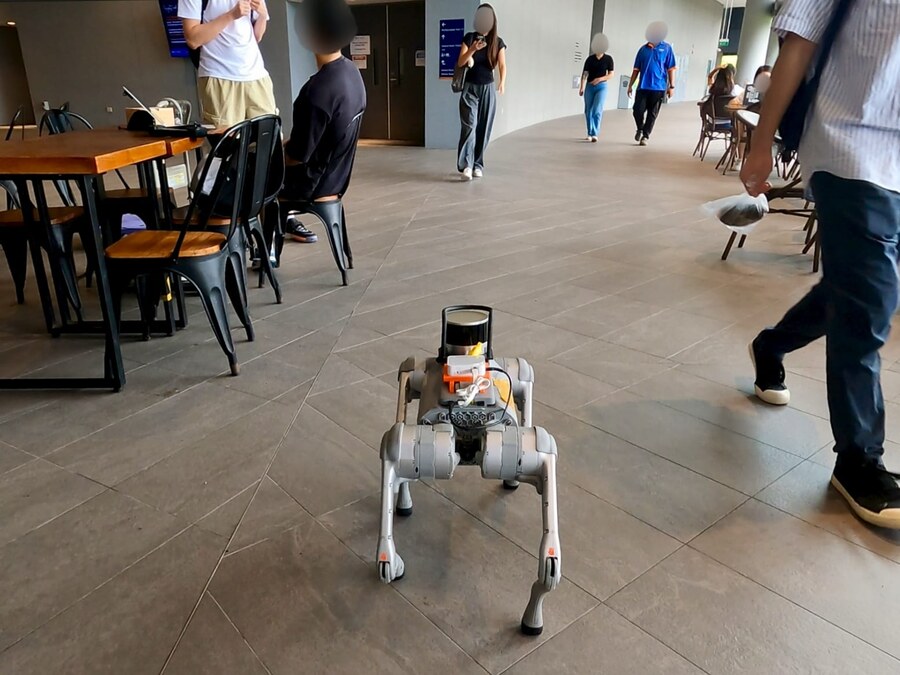}\hfill
\includegraphics[height=2.1cm]{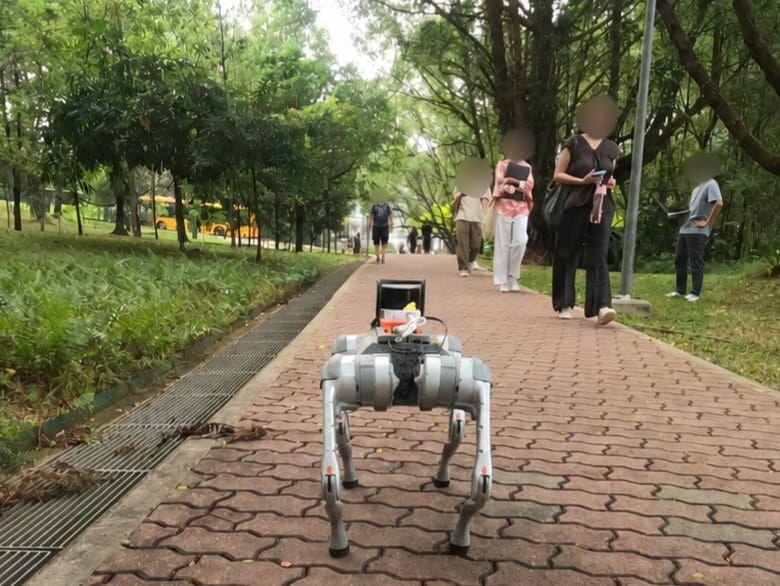}
\caption{(a) Unitree Go2 with RGB-D camera and LiDAR; (b,\,c) the indoor and outdoor test scenes.}
\label{fig:real}
\end{figure}

\begin{table}[t]
\caption{Real-robot results (10 trials per method and scene; means over successes).}
\label{tab:real}
\centering
\footnotesize
\setlength{\tabcolsep}{3pt}
\begin{tabular}{llccc}
\toprule
Scene & Method & Success$\uparrow$ & Mean time (s) & Mean path (m) \\
\midrule
\multirow{3}{*}{Indoor}  & Nav2-MPPI & 5/10  & 28.6 & 9.64 \\
                                           & \ours{} (Traj.\ peds) & 7/10  & 32.1 & 10.96 \\
                                           & \ours{}   & 9/10  & 31.1 & 10.14 \\
\midrule
\multirow{3}{*}{Outdoor} & Nav2-MPPI & 3/10  & 18.9 & 7.74 \\
                                           & \ours{} (Traj.\ peds) & 8/10  & 16.1 & 8.29 \\
                                           & \ours{}   & 10/10 & 17.9 & 8.59 \\
\bottomrule
\end{tabular}
\vspace{-8px}
\end{table}

\subsection{Real-robot deployment}
\label{sec:real}

\textbf{Setup.} We deploy the policies on a Unitree Go2 quadruped equipped with an RGB-D camera and LiDAR (Fig.~\ref{fig:real}a). The traversability state is constructed online from RGB-D and odometry, while pedestrians are detected and tracked from LiDAR with a YOLO11 prior. LiDAR points are clustered with DBSCAN without filtering them by traversability; a pedestrian track is initialized only when both YOLO and a corresponding LiDAR cluster confirm the detection. After initialization, the pedestrian is tracked from LiDAR using a Kalman filter, with Hungarian association based on Euclidean distance. Commands are scaled to the robot's speed limit, and no global map, planner, or policy fine-tuning is used. Because the policy only needs a relative goal, it can also serve as the local controller of a mapping system that supplies waypoints, such as the topological mapping system CROSS~\cite{wang2026cross}, for long-range navigation. 

\textbf{Baseline.} We compare against Nav2's MPPI controller~\cite{macenski2020nav2}. Since a geometric LiDAR costmap cannot distinguish traversable terrain such as pavement from grass, MPPI uses the same traversability memory as our policy, re-expressed as a graded costmap with LiDAR obstacles overlaid. All methods use the same velocity smoother and collision monitor. MPPI treats detected pedestrians as occupied costmap cells and therefore does not explicitly reason about their velocity, whereas both learned policies receive tracked pedestrian positions and velocities. We additionally evaluate the Trajectron++-pedestrian policy from Sec.~\ref{sec:ablation}, trained with reactive learned pedestrian dynamics instead of direct trajectory replay. 

\textbf{Protocol and results.} In one indoor and one outdoor scene (Fig.~\ref{fig:real}b,\,c), the robot navigates to a goal 8--10\,m away through five scripted encounters, each repeated twice per method. A trial is successful if the robot reaches the goal without static collision or coming within 0.3\,m of a pedestrian. The replay-trained policy succeeds in 19/20 trials, compared with 15/20 for the Trajectron++-trained variant and 8/20 for MPPI (Table~\ref{tab:real}). MPPI often maintains its course until a moving pedestrian enters the local costmap near the robot, causing late avoidance. In contrast, both learned policies use pedestrian velocity estimates to anticipate the encounter and can slow down or move aside before the crossing point. The replay-trained policy is the most reliable despite never using a reactive pedestrian model during training, consistent with the held-out-video and Arena results. The replay policy's single failure occurs indoors, where the 0.5\,m/s hardware speed limit, substantially below the speeds typically selected during training, leaves insufficient time to move out of an approaching person's path.

\section{Conclusion, Limitations, and Future Work}
\label{sec:limits}

In this work, we presented a scalable approach to learning social navigation from
monocular walking videos by constructing the simulator directly in the
policy's state space. Reconstructed traversability and replayed pedestrian
trajectories enable counterfactual actions and reinforcement learning
\textit{without} image rendering. Experiments demonstrate generalization to unseen
videos, transfer to an independent simulator with higher success and fewer
pedestrian collisions, and real-robot deployment without policy modification.

At present, our representation assumes approximately planar ground within each
$20\times20$\,m local window; substantial curvature or elevation changes
can distort the BEV projection. We also model motion only for pedestrians
and treat the remaining scene as static. This assumption suits pedestrian
paths, but moving vehicles and bicycles can introduce map errors and
contribute to failures in mixed-traffic scenes. Finally, pedestrian
states depend on human-pose detection and depth; detection errors and
sensor noise degrade velocity estimation and hence collision avoidance. These limitations are interesting avenues for future work.

\balance
{\footnotesize
\bibliographystyle{IEEEtran}
\bibliography{references}}

\end{document}

%% file: figures/teaser.tex
\begin{tikzpicture}[
  font=\sffamily\fontsize{8.5}{10}\selectfont,
  text=black!88,
  panel/.style={inner sep=0pt, outer sep=0pt, draw=black!25, line width=0.4pt},
  card/.style={draw=black!25, fill=white, rounded corners=2.5pt,
    inner sep=3pt, outer sep=0pt, align=center, line width=0.5pt},
  heading/.style={font=\sffamily\mdseries\fontsize{8.5}{10}\selectfont,
    align=center, inner sep=0pt},
  rowtitle/.style={font=\sffamily\mdseries\fontsize{9.5}{11}\selectfont,
    anchor=west, inner sep=0pt},
  note/.style={align=center, inner sep=0pt},
  icons/.style={font=\sffamily\fontsize{12}{13}\selectfont},
  arr/.style={-{Latex[length=1.6mm,width=1.3mm]}, line width=0.8pt, draw=black!65},
]
\definecolor{teabg}{HTML}{F1F6F7}
\definecolor{teaped}{HTML}{1F4FD8}
\definecolor{teacam}{HTML}{E0129B}
\definecolor{teatrav}{HTML}{20B520}
\definecolor{teablock}{HTML}{D92B23}
\path[use as bounding box] (0,0.08) rectangle (8.55,-7.58);

\fill[black!3, rounded corners=3pt] (0,0.06) rectangle (8.55,-2.26);
\node[rowtitle] at (0.15,-0.20) {Simulation in pixel space};

\node[card, minimum width=2.05cm, minimum height=1.27cm] (src) at (1.175,-1.13) {};
\node[icons] at (1.175,-0.82) {\faBrain\hspace{4pt}\faPencilRuler\hspace{4pt}\faVideo};
\node[note] at (1.175,-1.38) {LLM / designer\\or video};
\node[heading] at (1.175,-2.01) {Assets / layout};

\node[card, minimum width=2.15cm, minimum height=1.27cm] (syn) at (3.925,-1.13) {};
\node[heading] at (3.925,-0.73) {3D scene};
\node[icons] at (3.925,-1.13) {\faCubes\quad\faWalking\,\faWalking};
\node[note] at (3.925,-1.51) {Scripted people};
\node[note] at (3.925,-2.01) {Human model};

\node[panel] (pixels) at (7.025,-1.13)
  {\includegraphics[trim=0 370bp 0 115bp,clip,height=1.27cm]{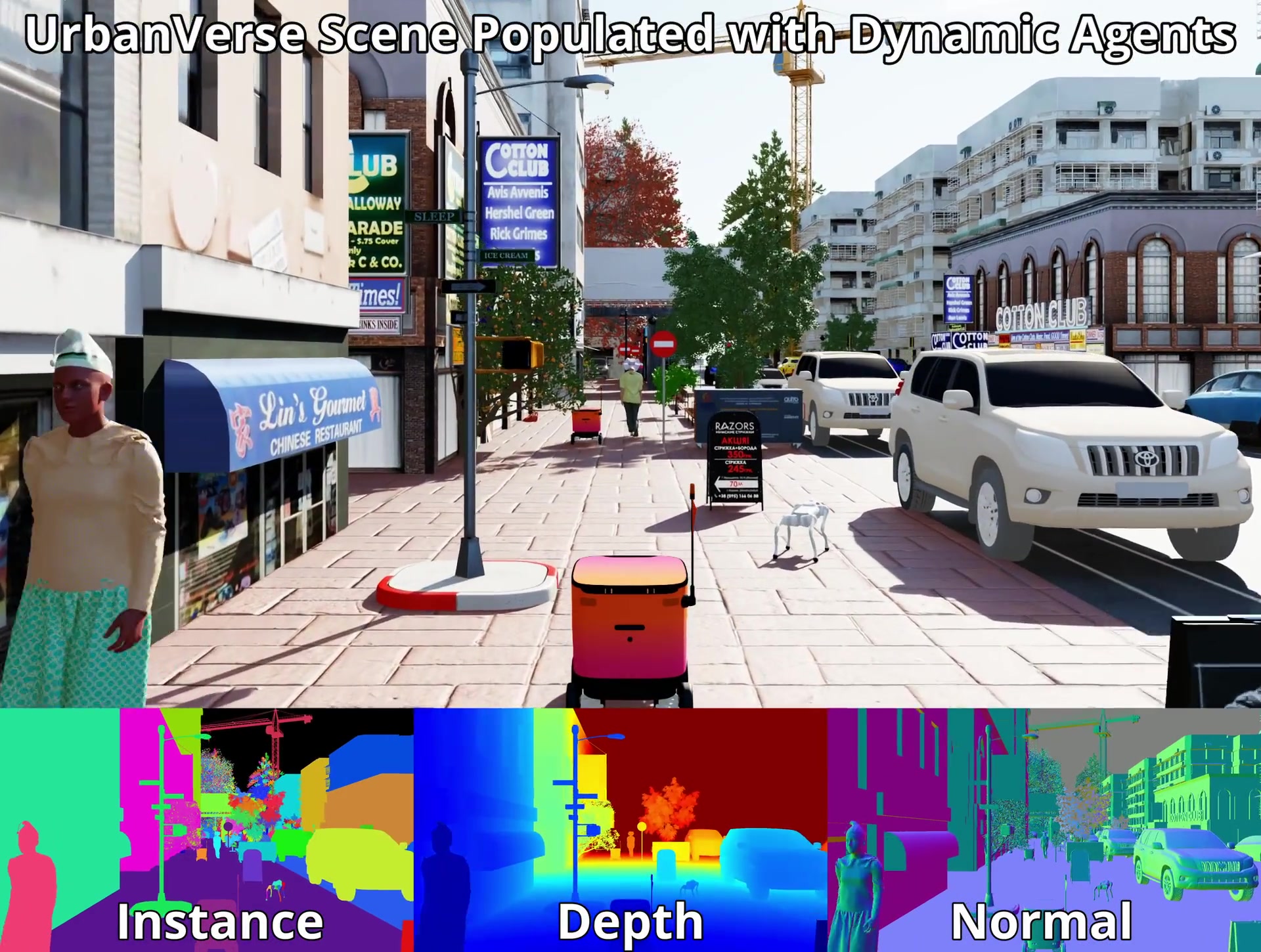}};
\node[heading] at (7.025,-2.01) {Policy on pixels};
\draw[arr] (src.east) -- (syn.west);
\draw[arr] (syn.east) -- (pixels.west);

\fill[teabg, rounded corners=3pt] (0,-2.44) rectangle (8.55,-6.75);
\node[rowtitle] at (0.15,-2.70) {Simulation in state space};

\node[heading] at (1.175,-3.80) {Walking video};
\node[panel, anchor=north west] at (0.25,-4.08)
  {\includegraphics[width=2.05cm]{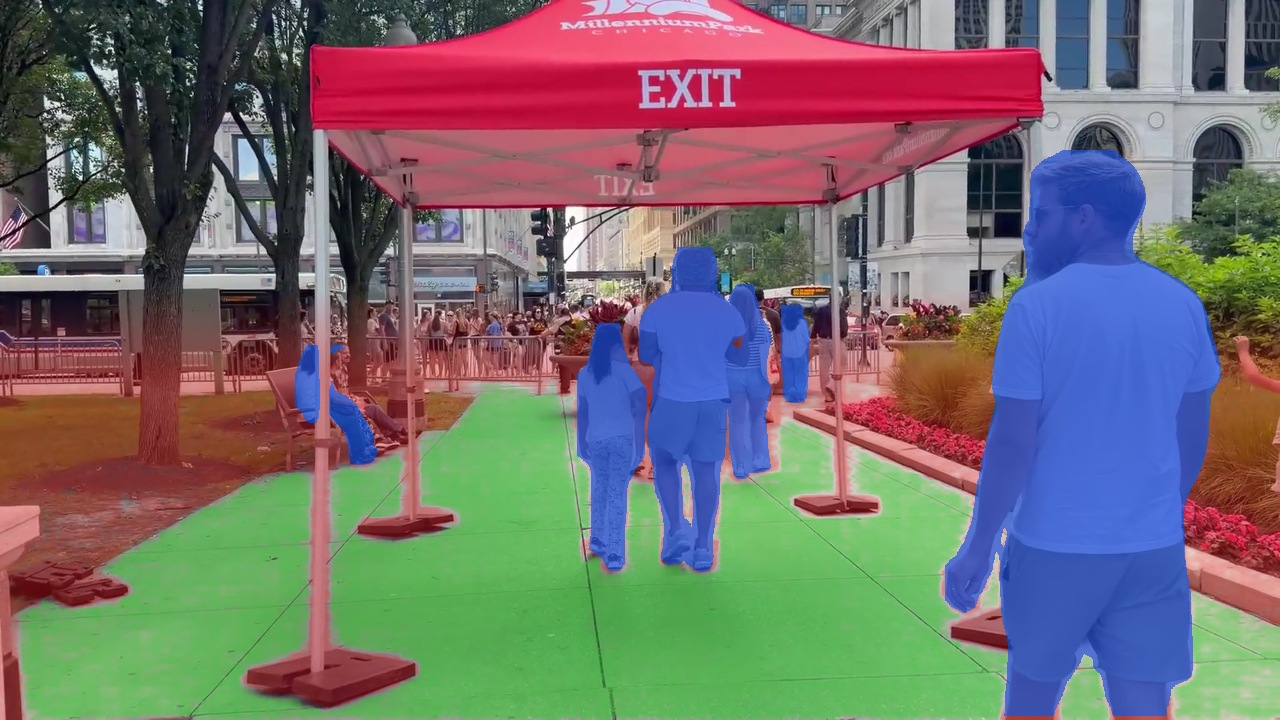}};
\node[panel, anchor=north west] at (0.20,-4.16)
  {\includegraphics[width=2.05cm]{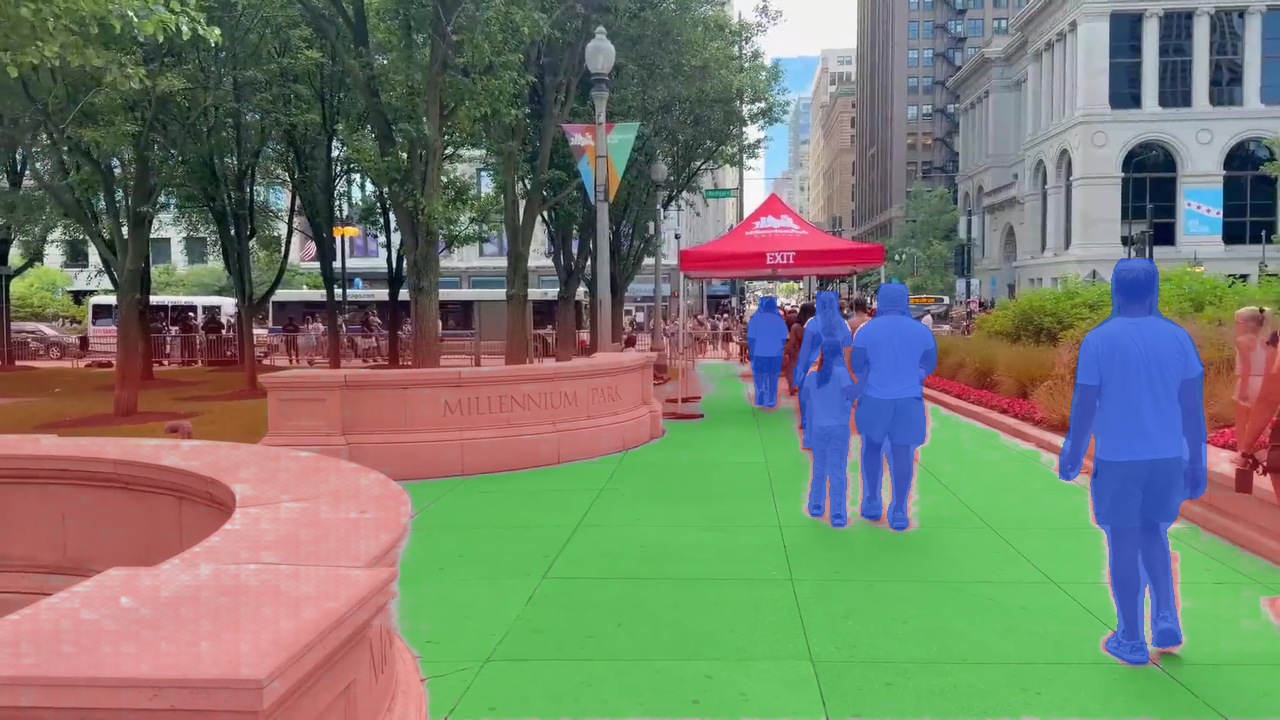}};
\node[panel, anchor=north west] (vid) at (0.15,-4.24)
  {\includegraphics[width=2.05cm]{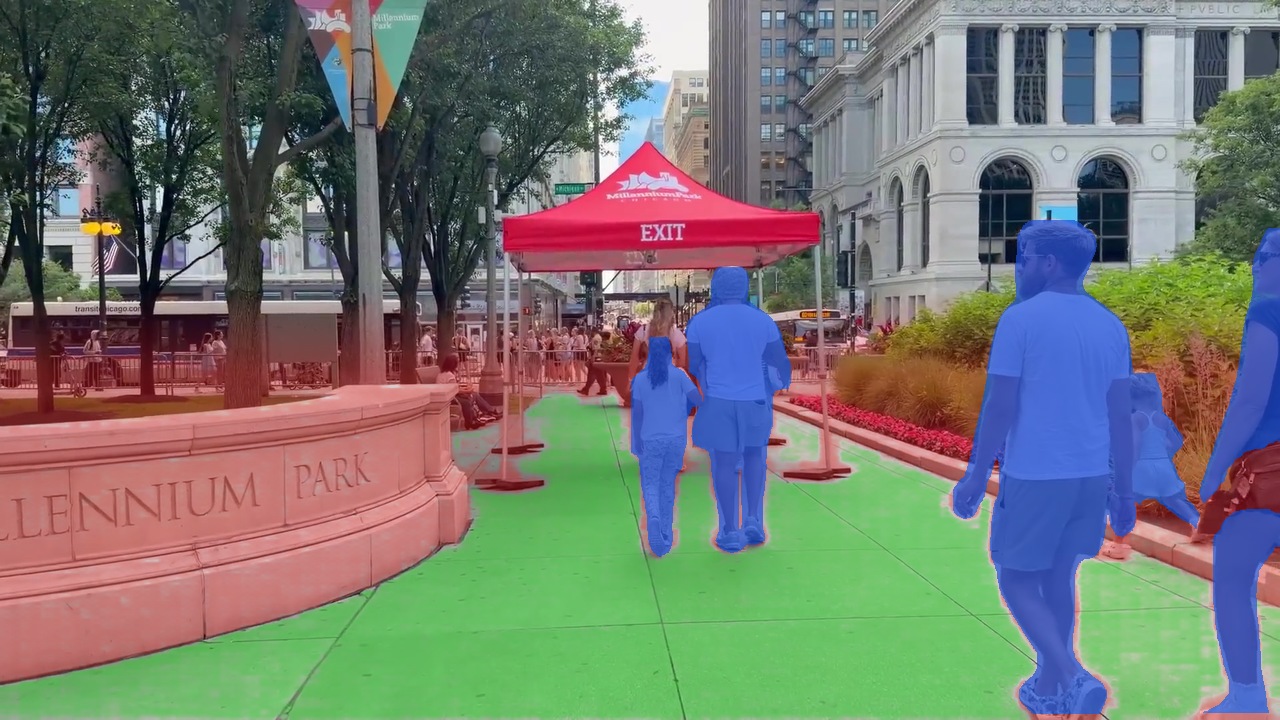}};
\node[note] at (1.175,-6.05) {Diverse real\\terrain\\+ real motion};

\node[heading] at (3.925,-3.08) {Static map};
\node[panel, anchor=north west] (st) at (2.85,-3.26)
  {\includegraphics[width=2.15cm]{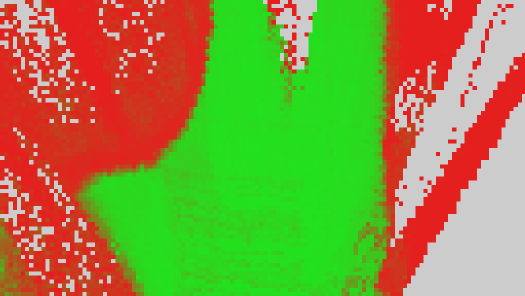}};
\node[note] at (3.925,-4.65) {Rigid warp};
\node[heading] at (3.925,-4.98) {Recorded people};
\node[panel, anchor=north west] (dy) at (2.85,-5.16)
  {\includegraphics[width=2.15cm]{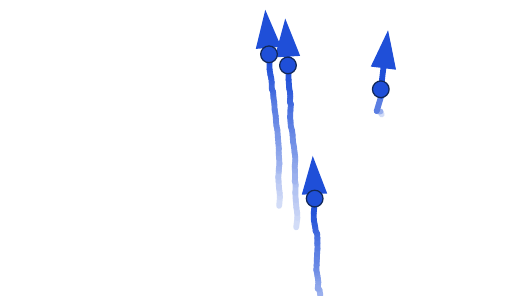}};
\node[note] at (3.925,-6.55) {Replay at time $t$};

\node[heading] at (7.025,-3.20) {Robot-frame state};
\node[panel, anchor=north west] (state) at (5.65,-3.44)
  {\includegraphics[width=2.75cm]{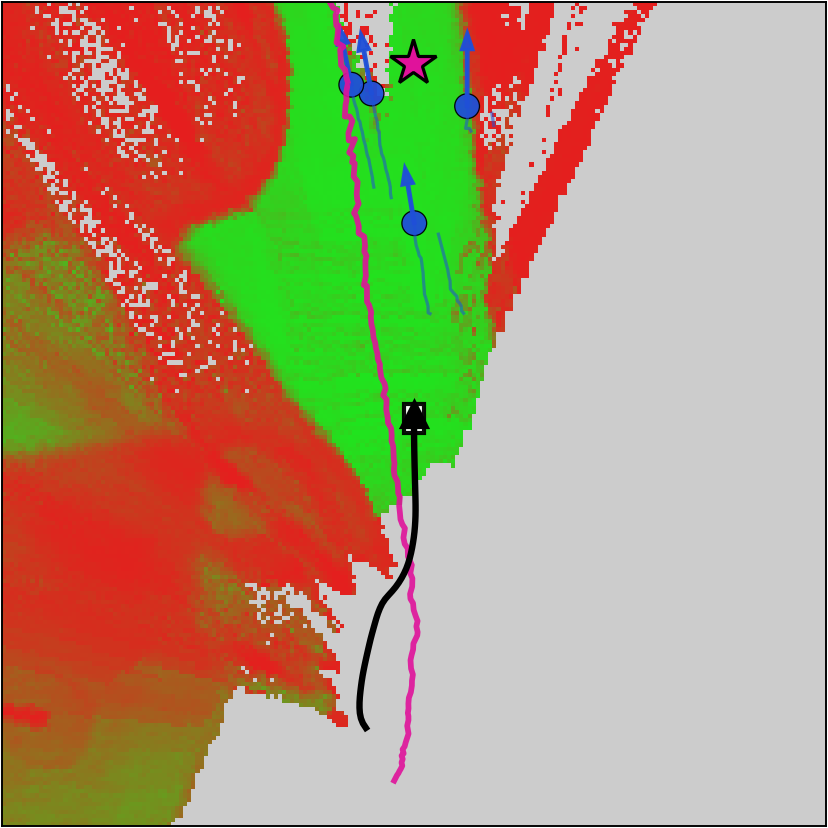}};

\coordinate (fork) at (2.48,0);
\draw[black!65, line width=0.8pt] (vid.east) -- (fork |- vid.east);
\draw[arr] (fork |- vid.east) |- (st.west);
\draw[arr] (fork |- vid.east) |- (dy.west);
\draw[arr] (st.east) -- (st.east -| state.west);
\draw[arr] (dy.east) -- (dy.east -| state.west);

\node[note, anchor=west] at (0.15,-7.02) {%
  {\color{teatrav}\rule[0.3ex]{0.22cm}{2.5pt}} traversable\quad
  {\color{teablock}\rule[0.3ex]{0.22cm}{2.5pt}} non-traversable\quad
  {\color{black!22}\rule[0.3ex]{0.22cm}{2.5pt}} unobserved};
\node[note, anchor=west] at (0.15,-7.43) {%
  {\color{teaped}\rule[0.3ex]{0.22cm}{1.5pt}} people\quad
  {\color{teacam}\rule[0.3ex]{0.22cm}{1.5pt}} camera path\quad
  {\color{black}\rule[0.3ex]{0.22cm}{1.5pt}} robot\quad
  {\color{teacam}$\bigstar$} goal};
\end{tikzpicture}